\documentclass[10pt,twocolumn,letterpaper]{article}

\usepackage{iccv}
\usepackage{times}
\usepackage{epsfig}
\usepackage{graphicx}
\usepackage{amsmath}
\usepackage{amssymb}
\usepackage[dvipsnames]{xcolor}
\usepackage{diagbox}
\usepackage{subcaption}
\usepackage[breaklinks=true,bookmarks=false]{hyperref}

\iccvfinalcopy

\def\iccvPaperID{912}

\ificcvfinal\fi

\newcommand{\smallsec}[1]{\vspace{0.01in} \noindent {\bf #1.}}

\newcommand\pred[1]{\texttt{#1}}
\newcommand\lowcorr[1]{{\color{ForestGreen}{\bf #1}}}
\newcommand\highcorr[1]{{\color{BrickRed}{#1}}}

\begin{document}

\title{SpatialSense: An Adversarially Crowdsourced Benchmark for\\Spatial Relation Recognition}

\author{Kaiyu Yang\\
Princeton University\\
{\tt\small kaiyuy@cs.princeton.edu}
\and
Olga Russakovsky \\
Princeton University\\
{\tt\small olgarus@cs.princeton.edu}
\and
Jia Deng\\
Princeton University\\
{\tt\small jiadeng@cs.princeton.edu}
}

\maketitle

\begin{abstract}
Understanding the spatial relations between objects in images is a surprisingly challenging task (Fig.~\ref{fig:samples}). A chair may be ``behind'' a person even if it appears to the left of the person in the image (depending on which way the person is facing). Two students that appear close to each other in the image may not in fact be ``next to'' each other if there is a third student between them.

We introduce SpatialSense, a dataset specializing in spatial relation recognition which captures a broad spectrum of such challenges, allowing for proper benchmarking of computer vision techniques. 
SpatialSense is constructed through adversarial crowdsourcing, in which human annotators are tasked with finding spatial relations that are difficult to predict using simple cues such as 2D spatial configuration or language priors.
Adversarial crowdsourcing significantly reduces dataset bias and samples more interesting relations in the long tail compared to existing datasets. 
On SpatialSense, state-of-the-art recognition models perform comparably to simple baselines, suggesting that they rely on straightforward cues instead of fully reasoning about this complex task. 
The SpatialSense benchmark provides a path forward to advancing the spatial reasoning capabilities of computer vision systems.
The dataset and code are available at \url{https://github.com/princeton-vl/SpatialSense}.
\end{abstract}
\vspace{-2mm} 

\section{Introduction}

Visual understanding of space is essential for an intelligent agent. Such an
understanding is the basis for describing scenes and referring to objects~\cite{guadarrama2013grounding}; it is also the
foundation required for tasks such as navigation and manipulation~\cite{zeng2018semantic}. To understand space it is important to understand \emph{spatial relations}, that is, how different spatial entities are
configured relative to each other to compose a scene. Consider the following description: ``Inside the living room, under
the window next to the wall is a table, on top of which lies a vase with flowers''. The sentence may be structured in many different ways, but its meaning is determined by the 
objects (room, window, table, vase, flowers) and their spatial relations (table in
room, table under window, table next to wall, vase on table, flowers in
vase). 

\begin{figure}[t]
\begin{center}
   \includegraphics[width=1.0\linewidth]{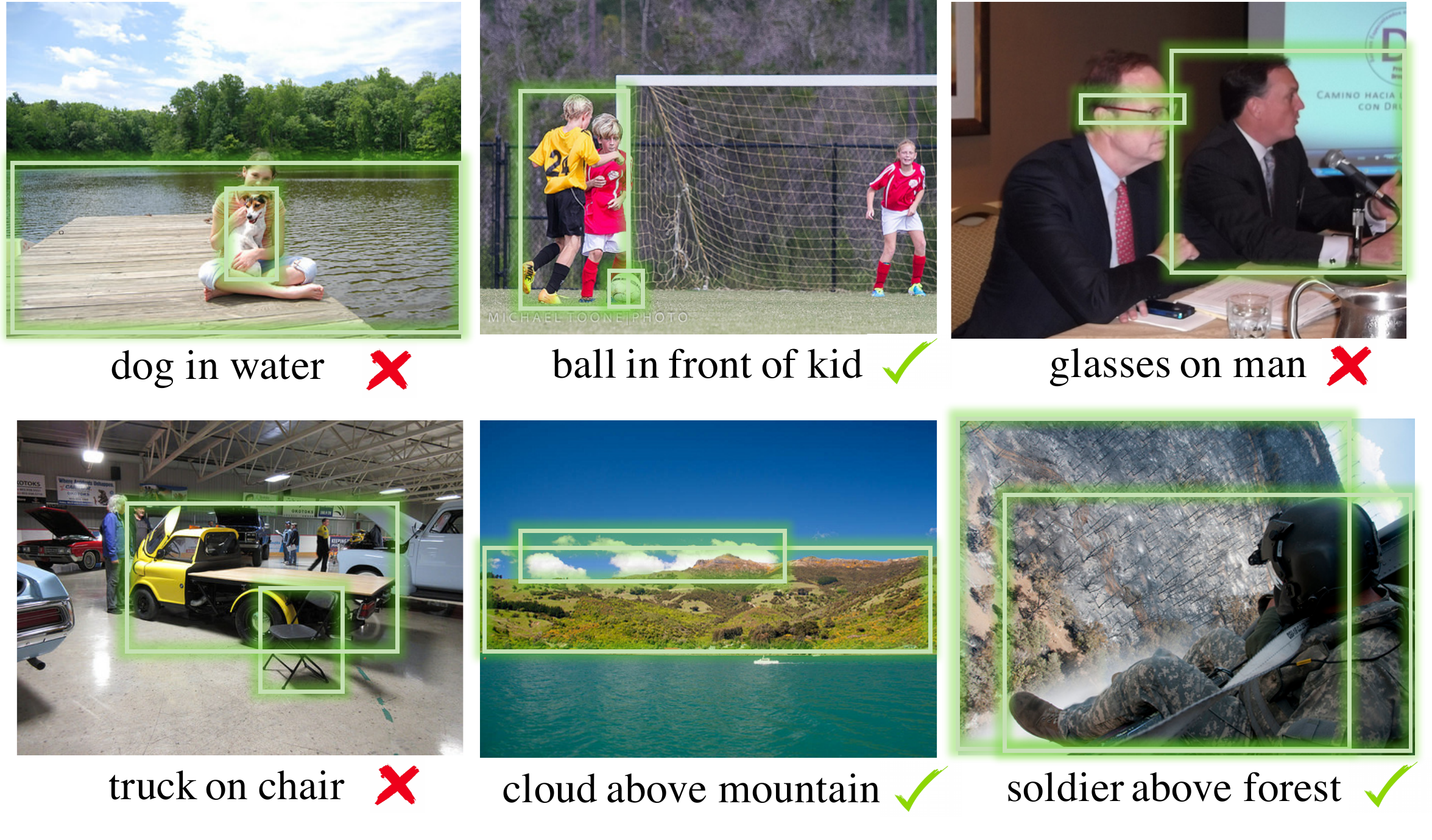}
\end{center}
\vspace{-8mm} 
\caption{Spatial relation recognition in images is a challenging task which requires a deep understanding of all the objects in the image, their 3D configuration, and their interactions. Understanding that the dog is \emph{not} in the water (top left) requires reasoning about the pier in addition to the dog and the water. Understanding that the ball is in fact in front of the kid despite being off to the right in the image space (top center) requires inferring the 3D spatial configuration. Properly evaluating computer vision abilities on this task is difficult because it requires collecting a benchmark covering the full spectrum of such challenges. We introduce SpatialSense, a novel dataset collected through \emph{adversarial crowdsourcing}, which provides a diverse and challenging testbed for the task of spatial relation recognition. 
}
\vspace{-4mm}
\label{fig:samples}
\end{figure}

This raises the problem of spatial relation recognition: given two objects in a scene, what is their spatial relation? This problem is important, interesting, and challenging because the semantics of spatial relations are rich and complex. The spatial semantics between objects depend not only on geometric properties such as location, pose, and shape, but also the frame of reference (\eg ``left of the car'' can be relative to the observer or the car) and object-specific common sense knowledge (\eg ``hand over bed'' does not imply physical contact while ``blanket over bed'' does).

\smallsec{Benchmarking spatial relations} Despite the importance of this problem, there is no benchmark dataset specializing
in spatial relations. Large datasets such as  
 Open Images~\cite{OpenImages}, Visual Genome~\cite{krishna2016visual}, and Visual Relationship Detection~\cite{lu2016visual} provide annotations for generic visual relations (human-verified \emph{subject-predicate-object} triplets such as ``person-riding-bike'') and include a significant number of spatial relations (38.2\%, 51.5\%, and 66.0\% respectively). But several characteristics make them less suitable for evaluating spatial relation recognition.

One issue in current visual relation datasets is significant language bias---the examples
are dominated by relations that can be guessed without an actual spatial understanding. For example, 66\% of all spatial relations in Visual Genome consist of one object ``on'' another. Among relations involving a table, 89.37\% of them define an object ``on'' the table. This means that a system can take advantage of such priors to do well without even looking at the image. This is undesirable because evaluating on these examples will not provide a proper gauge of an algorithm's ability to visually understand spatial relations. 

The second issue is in the evaluation metric. As collecting exhaustive relation annotations in an image is very challenging, metrics such as Recall@K (the recall of ground truth relations given K predicted relations) have been used~\cite{lu2016visual,Zhang_2017_CVPR,li2017vip,dai2017detecting,Zhang_2017_ICCV,Peyre_2017_ICCV}. However, this metric fails to distinguish between a good system producing valid albeit unannotated predictions and a bad system producing false positives, as well as fails to fully evaluate the system's ability to distinguish between positive and negative relations. 


\smallsec{Contributions} In this paper we introduce \emph{SpatialSense}, a dataset for spatial
relation recognition. A key feature of the dataset is that it is constructed through \emph{adversarial crowdsourcing}: a human annotator is asked to identify adversarial examples to confuse a robot. This ensures that the dataset focuses on questions that require more advanced reasoning and cannot be answered by simple spatial and language priors. Each annotator is explicitly tasked with identifying either positive or negative relations, ensuring an equal representation of each within the dataset. To avoid the problems arising out of non-exhaustive annotations, we formulate the task as a binary classification of individual relations. 


SpatialSense has 17,498 relations on 11,569 images. 
Given two objects (names and bounding boxes), the task is to classify whether a particular spatial relation holds. We provide the object names and localizations to decouple object detection from spatial relation recognition, such that a successful relation recognition system can be directly placed on top of any object detection system.
The dataset contains relations between 3,679 unique object classes, with 2,139 of these object classes appearing only once, providing a challenging long-tail distribution of concepts.

SpatialSense provides a rigorous testbed for spatial relation reasoning that is not easily amenable to simple priors. First, each predicate (``on'', ``under'', etc.) has an equal number of positive and negative relations. Second, simple baselines using only 2D or language cues are significantly less effective on SpatialSense than on other existing spatial reasoning benchmarks.
SpatialSense is complementary to large-scale datasets such as Visual Genome or Open Images, in that it enables testing spatial relation recognition models with challenging examples in the long tail.

We evaluate multiple state-of-the-art visual relationship detection models on SpatialSense. Experimental results reveal that these models rely too much on dataset bias and now perform comparably to simple baselines. This demonstrates that adversarial crowdsourcing is effective for reducing dataset bias, and showcases that SpatialSense is an important step towards improving spatial reasoning capabilities of computer vision systems.

\section{Related Work}

\smallsec{Visual relationship recognition}
Recognition of visual relations has recently emerged as a frontier of high-level computer vision moving beyond object recognition.
Sadeghi \& Farhadi~\cite{sadeghi2011recognition} studied detecting \emph{visual phrases} from images. 
A visual phrase can be a spatial relation (\eg ``person next to bicycle'').  
But their dataset contains only 17 unique visual phrases, 9 of which are spatial relations. 
This means that each spatial predicate only occurs with a small number of object categories: 
\eg ``next to'' only occurs with ``person'',``car'', and ``bicycle''. 
Thus the dataset is unsuitable for evaluating a general understanding of ``next to'' that is agnostic to object categories.

Lu \etal~\cite{lu2016visual} introduced the task of \emph{visual relationship detection}
---given an image, the algorithm predicts \emph{subject-predicate-object} triplets as well as the object bounding boxes. 
In contrast, our task is classification rather than detection, the object pairs are given, and there are both positive and negative relations.
Our task setup leads to a proper evaluation of relation understanding.

The VRD dataset introduced by Lu \etal~\cite{lu2016visual} includes spatial relations, but unlike our dataset, does not have negative examples. 
Thus evaluation using VRD has been based on Recall@K, which is ill-suited for spatial relations
because numerous valid spatial relations can hold in an image, making it difficult to 
choose the appropriate K. 
In addition, as we will show in Section~\ref{section4comparisons}, the spatial relations in VRD are significantly easier to predict using simple priors. 
Visual Genome~\cite{krishna2016visual} is another dataset that is larger and has extensive annotations of visual relations. 
Similar to VRD, it covers a significant number of spatial relations but has no negative examples,
and the spatial relations are easily predictable from simple priors. 

VRD and Visual Genome have spurred the development of new approaches for visual relationship detection. 
Successful methods typically build on top of an object detection module, and reason jointly over language and visual features~\cite{li2017vip,Zhang_2017_CVPR,liang2017deep,dai2017detecting,Zhuang_2017_ICCV,Yu_2017_ICCV}.
Multiple independent directions have been proven fruitful, including 
learning features that are agnostic to object categories~\cite{yang2018shuffle},
facilitating the interaction between object features and predicate features~\cite{yang2018shuffle,dai2017detecting,li2017vip}, overcoming the scarcity of labeled data through weakly supervised learning~\cite{Peyre_2017_ICCV,Zhang_2017_ICCV}, 
and detecting the relationships among multiple objects jointly as scene graphs~\cite{xu2017scene, li2017scene, woo2018linknet, li2018factorizable}.
We benchmark some of the state-of-the-art approaches on our dataset and compare them with simple baselines based on language or 2D cues.

Peyre \etal introduced a dataset of unusual relations (UnRel)~\cite{Peyre_2017_ICCV} sharing similar motivation to ours.
To address the problem of missing annotations, the relations in UnRel are annotated exhaustively. 
Annotating every instance is made feasible by having a small predefined list of relations that are carefully designed 
to be unusual (such as ``car under elephant''). 
This method is not scalable to a large number of relations.
First, it is difficult to manually pick a large set of unusual relations.
Second, for annotating every single instance, the amount of crowdsourcing efforts grows linearly with the number of relations.
Our method samples interesting relations by encouraging crowd workers to discover them in images, and thus circumvents the scalability problem.
As a result, UnRel has 76 unique relation triplets formed by 18 predicates while SpatialSense has 13,229 unique relations with 9 predicates.

Open Images~\cite{OpenImages} is a recent large-scale dataset containing a significant number of visual relations. Two of its design choices mitigate the problem of non-exhaustive annotation and the resulting ill-posed evaluation metric: (1) the inclusion of negative relations makes it possible to identify some false positives in detected relations, and (2) the relation annotations are dense in the sense that for every \emph{annotated} image-level label (e.g., a bottle is present in the image), humans have drawn bounding boxes around all object instances of that class, and have verified the presence of each relation in a predefined relation vocabulary. Therefore, Open Images enables more accurate evaluation of visual relationship detection methods. However, since the portion of annotated image-level labels in Open Images is approximately 43\%~\cite{OpenImages}, it remains impossible to evaluate any detected relation between objects in the remaining 57\%.

\smallsec{Dataset bias}
Instead of studying the misaligned distributions between data and the visual world~\cite{torralba2011unbiased}, we focus on a specific aspect of dataset bias, which allows models to take shortcuts leading to a superficial impression of good performance.
Extensive research on such bias has been conducted in the context of visual question answering (VQA)~\cite{zhang2016yin,goyal2016making,chao2017being,kafle2017analysis,agrawal2018don,manjunatha2019explicit}.
Many VQA datasets suffer from language bias;
the questions can be answered well simply by using language priors while ignoring images.
Zhang \etal~\cite{zhang2016yin} balance the data for yes/no questions on abstract scenes.
They show a question-image pair and ask the annotator to compose a new scene on which the answer for the question is different.
Goyal \etal~\cite{goyal2016making} applied the same idea to real images.
Instead of asking annotators to create new scenes, they provide a few semantically similar images for the annotator to choose from.

Our adversarial crowdsourcing approach addresses the same issue (ensuring the input image is required to answer questions) 
in a substantially different way.  
Spatial relation recognition can be understood as a special case of VQA, where the
questions are restricted to verifying spatial relations. In this sense, 
Zhang \etal~\cite{zhang2016yin} and Goyal \etal~\cite{goyal2016making} ask the crowd to
select hard images---images that defy the expected answers from language priors---with the
questions fixed, whereas we ask the crowd to select hard questions with
the images fixed. One potential advantage of selecting hard questions is that humans
can easily compose new questions but cannot easily synthesize photorealistic images, and it
can also be hard to find images that defy language priors---those images are by
definition less common because language priors reflect common occurrences.

\smallsec{Adversarial crowdsourcing} Our adversarial crowdsourcing approach is inspired by
the ``Beat the Machine'' framework~\cite{attenberg2011beat}, in which
a worker is challenged to find cases that will cause an AI system to
fail. Adversarial crowdsourcing is related to active learning (\eg \cite{kapoor2007active,vijayanarasimhan2014large}) in that in both cases we
seek difficult examples to improve learning. The key difference, however, is that in active
learning it is the machine's task to identify hard examples whereas in adversarial
crowdsourcing it is on the human annotator.

\section{Dataset Collection through Adversarial Crowdsourcing}

\label{section3method}

\begin{figure*}[t]
\begin{center}
   \vspace{-7mm}
   \includegraphics[width=0.9\linewidth]{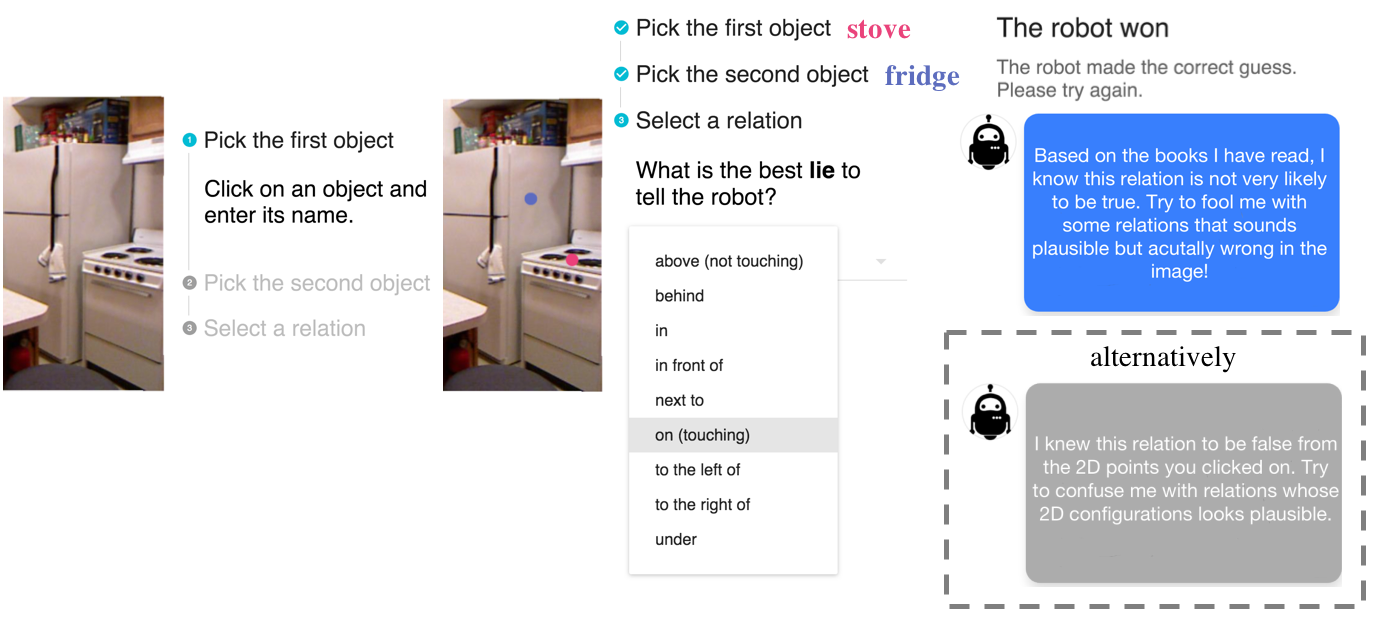}
\end{center}
\vspace{-8mm}
   \caption{When collecting negative examples, the annotator picks a pair of objects and lies about their spatial relation (``stove on fridge'') with the goal of making it believable enough for the robot. However, here the robot catches the annotator in the lie and explains how the correct guess was made (from language or 2D cues, or both). The UI for positive examples is similar, in which the annotator tells the robot a spatial relation that is valid but unbelievable.}
\vspace{-4mm}
\label{fig:ui}
\end{figure*}

Datasets are meant to evaluate the performance of algorithms under challenging and varied conditions. However, one weakness observed in many datasets is a strong language bias, allowing algorithms to perform well by exploiting language priors even while ignoring the visual input~\cite{zhang2016yin,goyal2016making,chao2017being,kafle2017analysis,agrawal2018don}. Further, in the context of spatial reasoning, algorithms may exploit simple 2D cues, circumventing a true 3D understanding of spacing~\cite{johnson2016clevr}. We address both issues in our adversarial crowdsourcing framework.

\smallsec{Adversarial crowdsourcing protocol} In our data collection pipeline (Fig.~\ref{fig:ui}), we ask annotators to propose spatial relations to make a robot fail.
Given an image and a request for a positive or negative spatial relation, the annotator propose an example by clicking on two objects, entering their names and selecting a spatial predicate corresponding to the relation between them (a true relation if the request was for a positive one, and a false relation otherwise). The robot then tries to guess whether the relation is positive or negative using only the object names and the 2D coordinates given by the clicks. The task is completed if the robot is wrong, e.g., it predicts that the relation is positive but in fact it was negative. Otherwise, if the robot is able to guess correctly, the robot provides feedback to the annotator about how the correct guess was made, and the annotator tries again. Additional crowdsourcing is used to verify the collected relations and annotate the object bounding boxes.

To reduce language bias and promote true 3D spatial understanding, we need the annotators to pick relations that are difficult to predict given object names and 2D cues.
The robot is therefore an ensemble of two models: a language-only model and a 2D-only model. 
The language-only model takes two object names along with the predicate, and outputs the probability that the relation holds.
The object names are converted to word embeddings using Word2Vec~\cite{mikolov2013distributed}, which are then encoded into a fixed-length feature vector by a gated recurrent unit (GRU)~\cite{cho2014learning}.
The one hot encoding of the predicate is mapped to a vector of the same size by a linear layer.
The three feature vectors are fused by element-wise addition, on top of which a 2-layer fully connected network outputs the probability.
For the 2D-only model, linear layers map the object coordinates to feature vectors, and the prediction is made following the same procedure of the language-only model.
The final output of the robot is the average of these two models.
Initially, we trained the robot on a dataset of 7,850 relations collected without adversarial crowdsourcing; during the collection of SpatialSense, we occasionally re-trained the robot using all currently available data in order to prevent the annotators from exploiting the failing modes of a particular robot.

\smallsec{Concept vocabulary} We restrict the spatial predicate to a predefined list (\pred{above}, \pred{behind}, \pred{in}, \pred{in front of}, \pred{next to}, \pred{on}, \pred{to the left of}, \pred{to the right of}, \pred{under}) instead of letting human annotators enter free-form text. Spatial relations can be encoded using a surprisingly small set of prepositions~\cite{landau1993whence}. Our list of 9 predicates covers the coarse-grained semantics of most spatial relations. Although it is possible to extend the vocabulary to represent more fine-grained spatial semantics (such as \pred{lean on} and \pred{sit on}), fewer predicates ensures sufficient training samples for each predicate. 
Nevertheless, our adversarial crowdsourcing method is scalable to more
predicates as the effort grows linearly w.r.t. the number of relations, but is independent of the number of predicates.

In contrast, for object names we allow the human annotator to enter free-form text. The vocabulary of objects is vast and it would be cumbersome for the human annotator to choose from a long list of objects. In addition, a limited vocabulary of objects would restrict the annotator's ability to pick the objects that form interesting
spatial relations. 
In general, an image is rarely completely predictable, so there will be an unusual or surprising spatial relation that beats the robot's simple intuition. 
This setup thus provides an efficient way of obtaining spatial relations in the long tail.

\begin{figure*}[ht]
\begin{center}
   \vspace{-4mm}
   \includegraphics[width=0.96\linewidth]{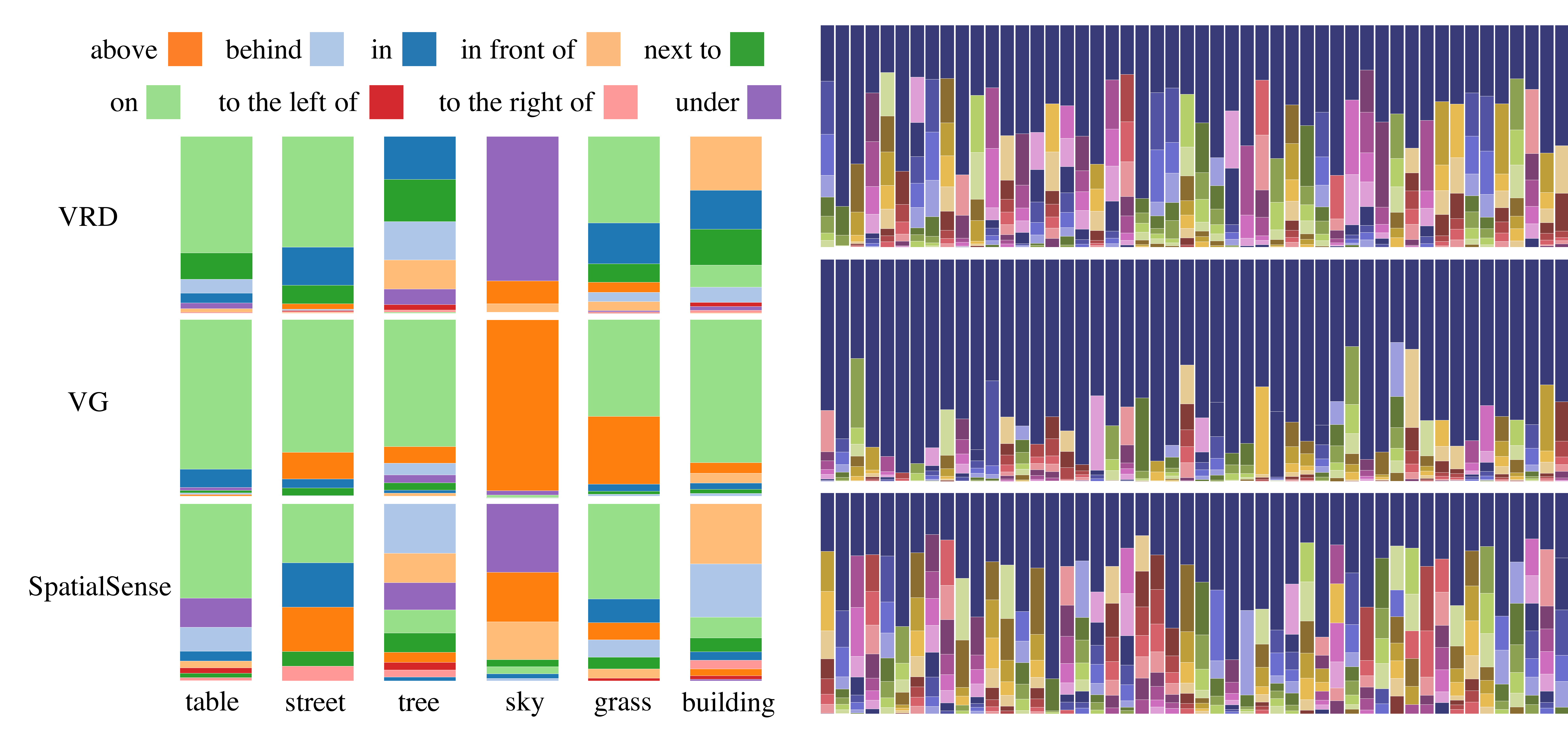}
\end{center}
\vspace{-8mm}
\caption{
(Left)
The predicate distributions of frequent objects in VRD-Spatial, VG-Spatial and SpatialSense-Positive.
For example, the bottom-left bar shows the frequency distribution of predicates \pred{on}, \pred{under}, \pred{behind}, etc. for the object ``table'' in SpatialSense. SpatialSense contains less language bias than other datasets since the distribution is more balanced. (Right) The predicate distributions of the top-50 objects in the three datasets, further showing the wider distribution in SpatialSense.
}
\vspace{-2mm}
\label{fig:compareObjects}
\end{figure*}

\smallsec{Image collection} We annotated $11,569$ images in total. Of these, 10,180 are RGB images from Flickr and $1,389$ are RGB-D images from NYU Depth~\cite{silberman2012indoor}  which we include to make it possible to test the utility of depth information for spatial understanding. When querying Flickr images, we use combinations of two keywords rather than a single
keyword, following the approach adopted by COCO~\cite{lin2014microsoft} to obtain images with diverse objects. Additionally, annotators can pick an image to annotate from a set of $8$ images, so as to avoid images that do not have enough objects (\eg a close up shot of a single foreground
object).  These techniques ensure that the images are complex scenes containing multiple objects necessary for relation reasoning.

We annotated $13,156$ relations on Flickr images and $4,342$ on NYU images. Each relation consists of a spatial predicate, the names of two objects, and their bounding boxes. Importantly, there is an equal number of positive \emph{and negative} relations for each of the 9 predicate. $20\%$ of the relations are reserved for testing and $15\%$ for validation.

\section{Analysis of the Dataset}

\label{section4comparisons}

The SpatialSense dataset has two key advantages compared to existing benchmarks. 
First, it contains positive as well as negative relations.
Second, it is constructed to be challenging; 
due to adversarial crowdsourcing, simple language and 2D priors are not enough to do well on this dataset.
We now perform an in-depth analysis, comparing SpatialSense to VRD~\cite{lu2016visual}, Visual Genome~\cite{krishna2016visual} and 
a version of itself without adversarial crowdsourcing.

\subsection{Comparison to Existing Datasets}
\label{subsec:section4comparisonDatasets}

\paragraph{Setup.} 
Since VRD and Visual Genome contain generic relations and no negative examples, we preprocess the data to allow for fair comparison:
(1) Only the positive examples in SpatialSense are considered; the resulting dataset is referred to as \emph{SpatialSense-Positive};
(2) We filter out non-spatial relations in VRD and Visual Genome; the resulting datasets are referred to as \emph{VRD-Spatial} and \emph{VG-Spatial}.
In addition to discarding non-spatial relations, we also map the
predicates in VRD and Visual Genome to their equivalents in our list of 9 spatial predicates.
For example, \pred{rest on}, \pred{park on} and \pred{lying on} in VRD are all mapped to \pred{on}.
For Visual Genome, since there is no closed vocabulary, we examined the top-100 most frequent predicates to figure out the mapping (\hyperref[appendix:mapping]{Appendix A}).

\smallsec{Predicate distribution} Compared to VRD and VG, the predicate distribution in SpatialSense is less biased. Fig.~\ref{fig:compareObjects} visualizes the distribution of predicates corresponding to different objects in the three datasets. For VG-Spatial, objects are frequently dominated by a single predicate, such as something \pred{on} table or something \pred{on} street. VRD-Spatial, which was annotated in-house rather than via crowdsourcing, looks more balanced; there are nevertheless a large number of objects \pred{on} street and or \pred{under} sky.
This confirms that many spatial relations in VRD and VG can be predicted without even looking at the image. In contrast, SpatialSense-Positive has a more balanced distributions, which reduces language bias, making it more difficult to guess the relation from the object names alone. There are plenty of unexpected or difficult to predict relations in any scene, and the key of our adversarial crowdsourcing approach is to encourage the annotators to reveal these.

\begin{figure}[t]
\begin{center}
   \vspace{-3mm}
   \includegraphics[width=0.95\linewidth]{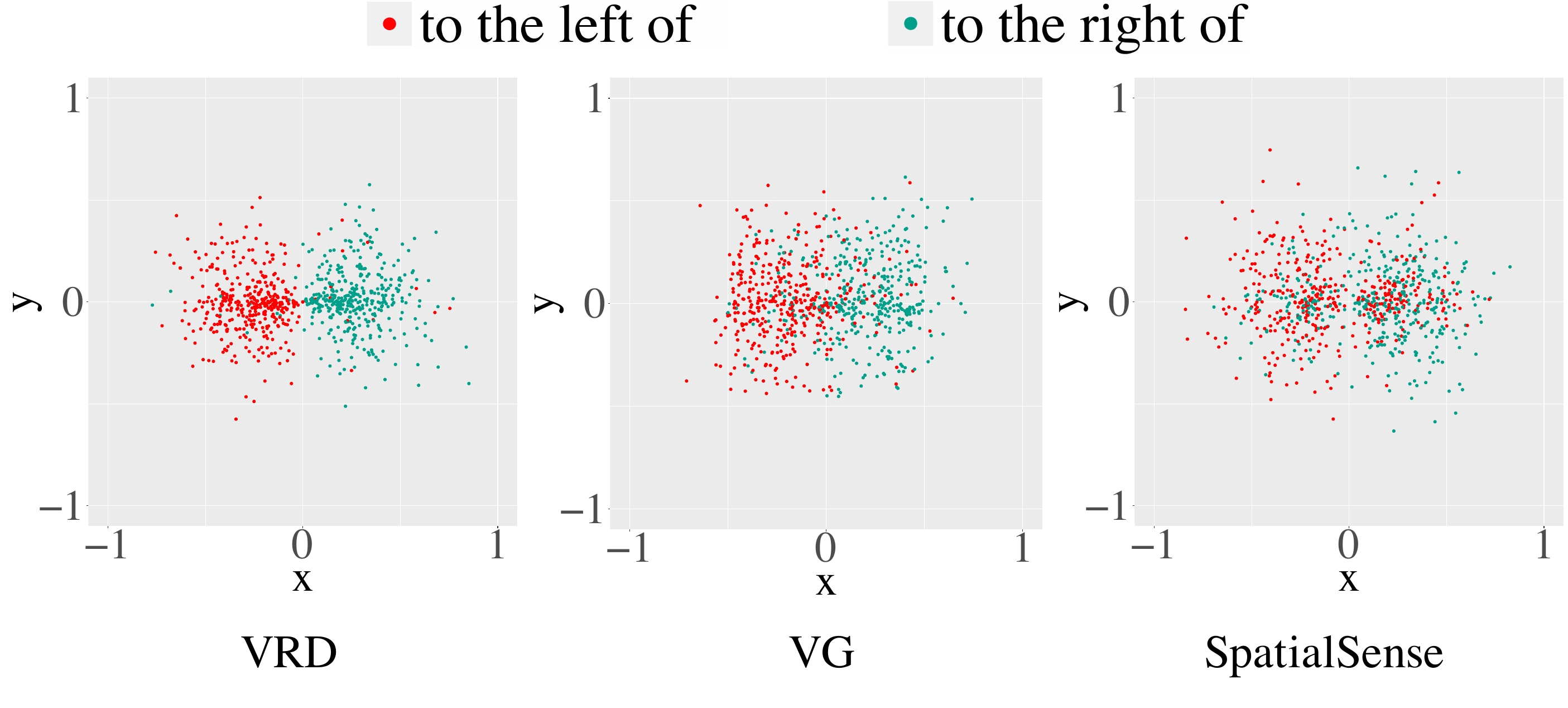}
\end{center}
\vspace{-8mm}
\caption{The 2D locations of subjects relative to objects for the predicates \pred{to the left/right of}, normalized by image size.
SpatialSense is less biased in 2D cues since the points are less separable.
Each figure contains 400 randomly sampled relations for each predicate.}
\vspace{-2mm}
\label{fig:2d}
\end{figure}

\smallsec{2D Spatial distribution} SpatialSense is also less biased in 2D cues. As an example, we evaluate whether the predicates \pred{to the left of} and \pred{to the right of} can be  predicted from spatial cues alone, without relying on pixel-level information. Concretely, for each subject-predicate-object relation, let $(x_s,y_s)$ be the center of the subject bounding box, $(x_o,y_o)$ be the center of the object bounding box, and $w,h$ be the image width and height. We compute the normalized relative location between the object and subject as $((x_s-x_o)/w,(y_s-y_o)/h)$. These are 2D points within $[-1, 1] \times [-1, 1]$ representing the 2D location of the subject relative to object. For VRD-Spatial and VG-Spatial, an algorithm can easily distinguish between the predicates \pred{to the left of} and \pred{to the right of} based on these 2D locations alone, as shown in Fig.~\ref{fig:2d}. For SpatialSense-Positive, however, these two predicates have similar distribution, making it difficult to tell them apart from 2D cues alone. Under our adversarial crowdsourcing framework, the annotators make extensive use of relative frame of references; a person standing to the left of a car can actually be on the right side of an image (when the car is facing towards the camera). Among other things, this relative frame of references makes SpatialSense less predictable from 2D cues and necessitates deeper spatial understanding.

\smallsec{Language and 2D baselines} 
To quantitatively show that SpatialSense is less biased in language and 2D locations, we examine the extent to which predicates can be determined from these cues alone, without pixel-level image information. For each relation, given the two object names (e.g., bed, floor) and their bounding boxes, a model has to predict the correct predicate (e.g., \pred{on}). We train and evaluate on the three datasets and compare the accuracies. For the comparison to be fair, VRD-Spatial and VG-Spatial are randomly sampled to have the same size as SpatialSense-Positive.  We adopt the official train/test split for VRD-Spatial and SpatialSense-Positive; for VG-Spatial, $20\%$ of the images are reserved for testing.

\begin{table}[t]
\resizebox{1.0\columnwidth}{!}{
\begin{tabular}{|l|ccc|c|}
\hline
\backslashbox{Train}{Test}  & VRD         & VG          & SpatialSense & average drop \\ \hline\hline
VRD                 & \textbf{66.9} / \textbf{59.6} & 45.2 / 53.2 & 30.6 / 39.9  & 29.0 / 13.1              \\
VG                  & 50.8 / 38.2 & \textbf{76.0} / \textbf{65.3} & 36.1 / 33.8  & 32.6 / 29.3              \\
SpatialSense        & 40.3 / 44.6 & 42.8 / 52.5 & \textbf{39.8} / \textbf{43.4}  & \textbf{-1.8} / \textbf{-5.2}  \\
\hline
\end{tabular}
}
\vspace{-3mm}
\caption{Accuracies of the language-only model / 2D-only model for predicting the spatial predicate from an object pair. Note two observations: (1) These simple baselines achieve low accuracy on SpatialSense compared to other datasets, demonstrating that SpatialSense is less susceptible to simple cues and requires more advanced reasoning. (2) SpatialSense is less biased than other datasets as evidenced by better cross-dataset generalization~\cite{torralba2011unbiased}. 
}
\vspace{-2mm}
\label{table:predicate_classification}
\end{table}

The model architectures are similar to those used in data collection:
For the language-only model, object names are encoded to fixed-length vectors using Word2Vec followed by a GRU.
The two vectors are fused into one by element-wise addition, which is then classified by a 2-layer fully connected network.
For the 2D-only model, bounding box coordinates are encoded by linear layers, and then fused and classified following the same procedure.
The models are trained with cross-entropy loss.

The results verify our intuition that SpatialSense is much more difficult to tackle using simple language and 2D cues than prior datasets (Table~\ref{table:predicate_classification}). Concretely, the language-only model achieves strong predicate prediction accuracy of $66.9\%$ on VRD-Spatial and $76.0\%$ on VG-Spatial, significantly higher than $39.8\%$ on SpatialSense-Positive. Similarly, the 2D-only baseline is able to achieve impressive accuracies of $59.6\%$ on VRD-Spatial and $65.3\%$ on VG-Spatial without ever seeing pixel information; in contrast, this simple model again struggles on SpatialSense-Positive and yields only $43.4\%$ accuracy. 

\smallsec{Cross-dataset generalization} Finally, we evaluate the generality of our collected dataset using the method proposed by Torralba and Efros~\cite{torralba2011unbiased}. Models trained on one dataset are evaluated on other datasets, and the resulting drop in accuracy is used as a metric for dataset bias (more drop corresponds to more bias). Table~\ref{table:predicate_classification} shows the results of cross-dataset predicate classification using the language-only and 2D-only model described above. Models trained on SpatialSense generalize impressively well to other datasets. The language-only model trained on SpatialSense but evaluated on VRD or VG (instead of on SpatialSense) achieves a $1.8\%$ average \emph{increase} in accuracy; similarly, the 2D model achieves a $5.2\%$ average \emph{increase} as well. In contrast, models trained on VRD or VG are not able to generalize well and experience a average $26.0\%$ \emph{drop} in accuracy when evaluated on a different dataset.

\subsection{The Effect of Adversarial Crowdsourcing} 

\label{comparewithnaive}

The bias reduction demonstrated in Section~\ref{subsec:section4comparisonDatasets} is a result of adversarial crowdsourcing. To verify, we compare with a dataset constructed without adversarial crowdsourcing. The dataset is collected by annotators who propose positive spatial relations freely (without the need to beat a robot), and the negative relations are randomly generated and verified by humans. The resulting dataset is SpatialNaive and contains 3,015 images with 3,925 positive and 3,925 negative relations. Just like in SpatialSense, each predicate has an equal number of positive and negative relations. 

We quantify the amount of bias in SpatialSense and SpatialNaive by comparing the performance on the task of spatial relation recognition.
Given two object names, their bounding boxes and a predicate, a model classifies whether or not the relation holds. Since SpatialNaive is smaller, we randomly sample a subset of SpatialSense, enforcing the two datasets to have exactly the same number of relations for each predicate. The model architectures are the same as those used for collecting data (described in section~\ref{section3method}), except that now the 2D locations are represented by object bounding boxes rather than coordinates. 20\% of the images in SpatialNaive are used for testing and another 15\% of the images are for hyperparameters tuning.

The results are in Table~\ref{table:compare_without_adversarial}. 
The models performs much worse on SpatialSense than SpatialNaive: $12.8\%$ accuracy drop for the language-only model, $6.1\%$ for the 2D-only, confirming the effectiveness of adversarial crowdsourcing for reducing dataset bias, especially the language bias.

\begin{table}
\vspace{-2mm}
\begin{center}
\begin{tabular}{|l|c|c|}
\hline
Dataset & Language & 2D Locations \\
\hline\hline
SpatialNaive &  69.2 & 71.3 \\
\hline
SpatialSense & $\mathbf{56.4}$ & $\mathbf{65.2}$ \\
\hline
\end{tabular}
\end{center}
\vspace{-6mm}
\caption{
SpatialSense, constructed with adversarial crowdsourcing, is significantly more challenging (lower accuracy of baselines) than the ablation dataset SpatialNaive.
}
\vspace{-3mm}
\label{table:compare_without_adversarial}
\end{table}

\begin{table*}[h]
\vspace{-4mm}
\centering
\resizebox{2.0\columnwidth}{!}{
\begin{tabular}{|l|c|ccccccccc|}
\hline
Model                             & Overall       & above         & behind        & in            & in front of   & next to          &    on             & to the left of & to the right of      & under   \\ \hline\hline
Language-only                     & 60.1          & 60.4          & 62.0          & 54.4          & 55.1          & 56.8             &   63.2           & 51.7           & 54.1                 & 70.3    \\ 
2D-only                           & 68.8          & 58.0          & 66.9          & \textbf{70.7} & 63.1          & 62.0             &  76.0           & 66.3           & \textbf{74.7}        & 67.9       \\ 
Language + 2D                     & 71.1 & 61.1          & 67.5          & 69.2          & 66.2          &  64.8    &  77.9           & \textbf{69.7}  & \textbf{74.7}        & \textbf{77.2}       \\ \hline
Vip-CNN~\cite{li2017vip}    & 67.2  & 55.6 & 68.1          & 66.0          & 62.7          & 62.3            & 72.5          & \textbf{69.7}           & 73.3              & 66.6       \\
Peyre \etal~\cite{Peyre_2017_ICCV}    & 67.5       & 59.0 & 67.1          & 69.8          & 57.8          & \textbf{65.7}             &  75.6           & 56.7           & 69.2                 & 66.2       \\ 
PPR-FCN~\cite{Zhuang_2017_ICCV} & 66.3  & 61.5 & 65.2          & 70.4          & 64.2          & 53.4            & 72.0          & 69.1           & 71.9              & 59.3         \\
DRNet~\cite{dai2017detecting}     & \textbf{71.3} & \textbf{62.8}          & \textbf{72.2} & 69.8          & \textbf{66.9} & 59.9             & \textbf{79.4}  & 63.5           & 66.4                 & 75.9      \\
VTransE~\cite{Zhang_2017_CVPR}    & 69.4          & 61.5 & 69.7          & 67.8          & 64.9          & 57.7             &  76.2           & 64.6           & 68.5                 & 76.9       \\ \hline
Human                             & 94.6          & 90.0         & 96.3         & 95.0          & 95.8         & 94.5             &  95.7           & 88.8           & 93.2                 & 94.1  \\ \hline
\end{tabular}
}
\vspace{-1mm}
\caption{The testing accuracies of baseline methods on spatial relation recognition. 
The 2D baseline performs closely to state-of-the-art models,
which suggests state-of-the-art models might learn to exploit simple priors and fail to develop deeper visual reasoning capabilities.}
\label{table:baselines-full}
\end{table*}

\section{Baselines for Spatial Relation Recognition}

Having verified that SpatialSense is an effective benchmark for spatial relation recognition, we evaluate multiple methods on SpatialSense, including simple baselines based on language and 2D cues as well as state-of-the-art models for visual relationship detection. Experimental results reveal the difficulty for state-of-the-art models to go beyond simple priors and learn to reason about visual content; a simple baseline based on 2D cues performs competitively with state-of-the-art models. We also conduct a human evaluation quantifying the level of ambiguity in the task.

\smallsec{Model architectures}
The task is spatial relation recognition: given the image, two objects (their names and bounding boxes) and a spatial predicate, the model classifies whether the relation holds.
Two simple baselines are evaluated: a language-only model and a 2D-only model.
Their architectures are the same as in section~\ref{comparewithnaive}.
We also report the performance of combining their predictions by a weighted average.
We evaluate five state-of-the-art models: Vip-CNN~\cite{li2017vip}, Peyre \etal~\cite{Peyre_2017_ICCV}, PPR-FCN~\cite{Zhuang_2017_ICCV}, DRNet~\cite{dai2017detecting} and VTransE~\cite{Zhang_2017_CVPR}.
They were created for visual relationship detection but can be adapted to our task straightforwardly:
First, object detectors are replaced by ground truth objects.
Second, object names are encoded using word embeddings rather than one hot encoding, since SpatialSense has unconstrained object categories. 
Third, for each relation \emph{subject-predicate-object}, the model takes \emph{subject} and \emph{object} as input, and generates scores for all predicates; 
the score for that particular \emph{predicate} is the final binary classification score.
Details are in \hyperref[appendix:architectures]{Appendix B}.

\smallsec{Implementation details}
Object names are encoded to fixed-length vectors using Word2Vec followed by a GRU.
When combining the language and 2D baselines by a weighted average, we find 80\% from 2D and 20\% from language to perform well (measured by validation accuracy).

For state-of-the-art models, we crop the union bounding box of the two objects, resize it to 280 $\times$ 280 and 
normalize the pixel values by the mean and standard deviation of all training images.
During training, we then crop it randomly, resize to 224 $\times$ 224 and apply color jittering;
during testing, we simply take a 224 $\times$ 224 crop at the center.

\smallsec{Analyzing the results} Table~\ref{table:baselines-full} summarizes the testing accuracies.
Due to the challenging nature of SpatialSense, the best models perform around 70\%, which is quite low for a binary classification task.
DRNet is the best model without ensemble, closely followed by other state-of-the-art models, which confirms that models for visual relationship detection can be well adapted to our task.
Notably, the 2D baseline performs closely to state-of-the-art models and is on a par with DRNet when combined with the language baseline.
This suggests state-of-the-art models might rely too much on 2D cues and fail to develop deeper visual reasoning.

\begin{table}
\vspace{-2mm}
\begin{center}
\resizebox{1.0\columnwidth}{!}{
\begin{tabular}{|l||cc|ccccc|}
\hline
 & L & \multicolumn{1}{c}{2D} & Vi & Pe & PP & D & VT\\
\hline\hline
Language-only & -- &  \lowcorr{0.04} & \lowcorr{0.05} & \lowcorr{0.09} & \lowcorr{0.05} & \highcorr{0.22}& \highcorr{0.33} \\
2D-only & \lowcorr{0.04}  &  -- & \highcorr{0.60} & \highcorr{0.35} & \highcorr{0.46} & \highcorr{0.43} & \highcorr{0.31} \\
\cline{2-8}
Vip-CNN~\cite{li2017vip} & \lowcorr{0.05}  & \highcorr{0.60} & -- &\highcorr{0.30}  & \highcorr{0.46} & \highcorr{0.25} & \highcorr{0.20} \\
Peyre \etal~\cite{Peyre_2017_ICCV} & \lowcorr{0.09}  & \highcorr{0.35} & \highcorr{0.30} & -- & \highcorr{0.30} & \highcorr{0.27}& \highcorr{0.20} \\
PPR-FCN~\cite{Zhuang_2017_ICCV}  &  \lowcorr{0.05}  & \highcorr{0.46} & \highcorr{0.46} & \highcorr{0.30} & -- &  \highcorr{0.24}& \highcorr{0.21} \\
DRNet~\cite{dai2017detecting} & \highcorr{0.22}  & \highcorr{0.43} & \highcorr{0.25} & \highcorr{0.27} & \highcorr{0.24} & --& \highcorr{0.44}\\
VTransE~\cite{Zhang_2017_CVPR} & \highcorr{0.33}  & \highcorr{0.31} & \highcorr{0.20} & \highcorr{0.20} & \highcorr{0.21} & \highcorr{0.44} & --\\
\hline
\end{tabular}
}
\end{center}
\vspace{-5mm}
\caption{
The correlation matrix between errors of the models, with low correlation in \lowcorr{green} and high correlation in \highcorr{red}. The 2D and language baselines make independent errors (low correlation). DRNet~\cite{dai2017detecting} and VTransE~\cite{Zhang_2017_CVPR} correlate strongly with both baselines, suggesting that their predictions are very similar to a combination of 2D and language cues. The other models are also highly correlated with the 2D baseline (they do not utilize language cues).
}
\label{table:errors_corr}
\end{table}

\begin{figure*}[t]
\vspace{-3mm}
\begin{center}
    \includegraphics[width=0.95\linewidth]{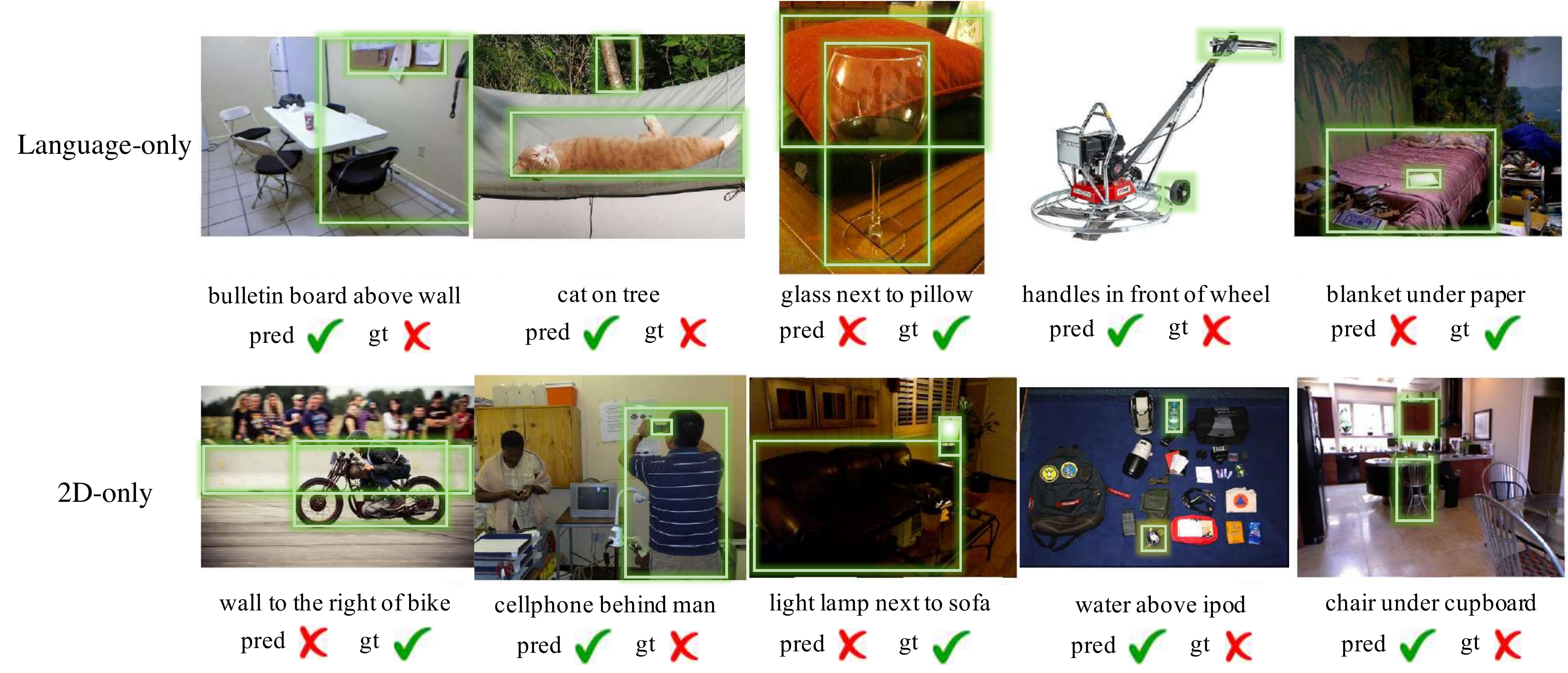}
\end{center}
\vspace{-5mm}
   \caption{Failing examples of the language and 2D baselines (\textit{pred}: prediction, \textit{gt}: ground truth).
   The language baseline fails when a frequent spatial relation does not occur in a particular image (e.g.,  ``cat on tree''), or a technically valid spatial relation is expressed in an unusual way (e.g., ``blanket under paper''). The 2D baseline fails to consider the relative frame of reference (e.g., ``wall to the right of bike'') and depth information (e.g., ``light lamp next to sofa'' and ``chair under cupboard'').
}
\label{fig:failing_examples}
\end{figure*}

To validate this conjecture, we examine the correlations between the errors.
For each model, consider an error vector of length N where N is the size of the testing data. The values in the vector are 1 when the model predicts incorrectly and 0 otherwise. Table~\ref{table:errors_corr} shows the correlation matrix between these error vectors.
The language baseline does not correlate well with the 2D baseline, suggesting that they make rather different kinds of errors.
However, all state-of-the-art models have high correlation with the 2D-only baseline, which implies they make similar predictions to a 2D-only model and supports our conjecture.

Failing examples of the simple baselines are in Fig.~\ref{fig:failing_examples}. Models based solely on 2D cues struggle in scenarios that involves the relative frame of reference or require depth-based reasoning. Language cues fall short when a common spatial relation does not appear or, in contrast, an unusual spatial relation is present. These observations indicate that neither language nor 2D cues are sufficient for spatial relation recognition. Beyond these simple cues, it is crucial to learn visual reasoning that elude the current state-of-the-art.
Our benchmark takes a step towards that goal by providing a more accurate gauge of a model's visual reasoning ability.

\smallsec{Human evaluation} Finally, recognizing spatial relations is inherently noisy; it is not always clear whether a relation holds.
We conduct a human evaluation, in which annotators are asked to make predictions on the testing data.
Multiple human responses on the same relation is merged by majority vote.
For quality control purpose, the annotators who answer ``yes'' more than 80\% of the time are considered outliers, and their responses are excluded. 
We collect 10,205 predictions and the accuracies are in the last row of Table~\ref{table:baselines-full}.
Although not perfect, humans perform very well on this task (94.6\%).
The gap between humans and algorithms provides a large room for future improvement on this benchmark.

\section{Conclusion}

We introduced a novel dataset SpatialSense for the challenging task of spatial relation recognition. SpatialSense was constructed through adversarial crowdsourcing, which significantly reduced its dataset bias compared to alternative datasets. We evaluated multiple baselines on SpatialSense, demonstrating e.g., that a simple 2D baseline performs competitively with state-of-the-art models. This reveals that state-of-the-art models rely too much on dataset bias in existing benchmarks, validating the need for SpatialSense as a new challenging testbed for spatial relation recognition.

\smallsec{Acknowledgements} This work is partially supported by National Science Foundation under grant 1633157,  DARPA under grant FA8750-18-2-0019, and  IARPA under grant D17PC00343.

\clearpage
\onecolumn
\setcounter{table}{0}
\renewcommand{\thetable}{\Alph{table}}
\setcounter{figure}{0}
\renewcommand{\thefigure}{\Alph{figure}}

\appendix

\section*{Appendix A: Mapping the Predicates from VRD and VG to SpatialSense}
\label{appendix:mapping}

In section 4.1, in order to make the three datasets comparable, 
we map the spatial predicates in VRD and Visual Genome to their equivalents in SpatialSense.
Here we describe the detailed mapping in Table ~\ref{table:mapping}.

\begin{table}[h]
\resizebox{1.0\columnwidth}{!}{
\begin{tabular}{|p{2cm}|p{2cm}|p{2cm}|p{2cm}|p{2cm}|p{2cm}|p{2cm}|p{2cm}|p{2cm}|p{2cm}|}
\hline
 SpatialSense   & above                                                                                  & behind                                                                                                  & in                                                                                                                                                                       & in front of                                                                                                     & next to                                                                                                                                        & on                                                                                                                                                                                                                                                                           & to the left of                                                                                                                        & to the right of                                                                                                                        & under                                                                                                                         \\ \hline
VRD & above, over                                                                            & behind, stand behind, sit behind, park behind                                                           & in, inside                                                                                                                                                               & in the front of                                                                                                 & sleep next to, sit next to, stand next to, park next, walk next to, beside, walk beside, adjacent to                                           & on, on the top of, sit on, stand on, drive on, park on, lying on, lean on, sleep on, rest on, skate on                                                                                                                                                                       & on the left of                                                                                                                        & on the right of                                                                                                                        & under, stand under, sit under, below, beneath                                                                                 \\ \hline
VG  & above, above a, above an, above the, are above, are above a, is above, on top of, over & behind, are behind, are behind a, behind a, behind an, behind the, is behind, is behind the, on back of & in, are in, are in a, are in an, are in the, flying in, hanging in, in a, in an, in the, inside, inside of, is in, is in a, is in the, laying in, sitting in, walking in & in front of, are in front of, are in front of a, in front of a, in front of an, in front of the, is in front of & next to, are next to, are next to a, are next to the, beside, is next to, is next to the, next to a, next to an, next to the, standing next to & on, are on, are on a, are on an, are on the, growing on, hanging on, is on, is on a, is on the, laying on, lying on, on a, on a a, on an, on are, on front of, on the, painted on, parked on, printed on, sitting on, sitting on top of, standing on, walking on, written on & are left of, in left, in left side of, left of, left side of, on left, on left of, on left side of, to left, to left of, to left of a & are to right of, on right, on right of, on right side, on right side of, right of, right side of, to right, to right of, to right of a & under, are under, are under a, are under an, below, beneath, is under, is under the, under a, under an, under the, underneath \\ \hline
\end{tabular}
}
\vspace{2mm}
\caption{
We map the spatial predicates in VRD and Visual Genome to our predefined list of 9 predicates.
We mannually check all predicates in VRD to figure out the mapping.
For Visual Genome, since there is no closed vocabulary,
we examined the top-100 most frequent predicates.
}
\label{table:mapping}
\end{table}

\section*{Appendix B: Model Architectures}
\label{appendix:architectures}

\begin{figure*}[h]
\begin{center}
   \includegraphics[width=1.0\linewidth]{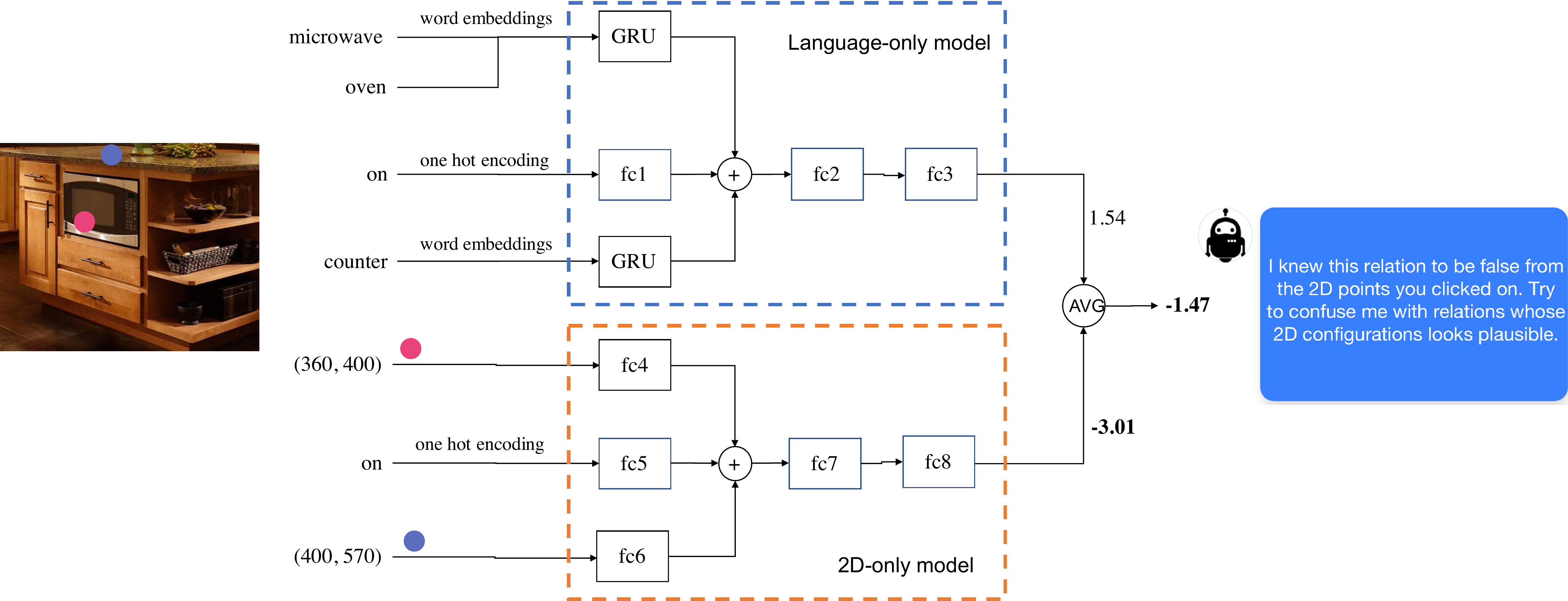}
\end{center}
   \caption{
In \textbf{adversarial crowdsourcing} (section 3 in our submission), the architecture of the robot is an ensemble of a language-only model and a 2D-only model. 
The language-only model takes two object names along with the predicate (``microwave oven'', ``on'', ``counter''), and outputs a score for the relation to hold (1.54).
The word embeddings of object names are encoded into 512-dimensional vectors by a gated recurrent unit (GRU)~\cite{cho2014learning} of 512 hidden units.
The same GRU is shared between the subject and the object.
The one hot encoding of the predicate is mapped to a 512-dimensional vector by a linear layer.
The three feature vectors are fused by element-wise addition, on top of which a 2-layer fully connected network (with 256 hidden units) outputs the score.
For the 2D-only model, linear layers map the object coordinates to 512-dimensional vectors, and others remain the same.
The final output is the average of these two models.
   }
\label{fig:robot}
\end{figure*}

\begin{figure*}[h]
\begin{center}
    \begin{subfigure}[t]{1.0\linewidth}
        \centering
        \includegraphics[width=1.0\linewidth]{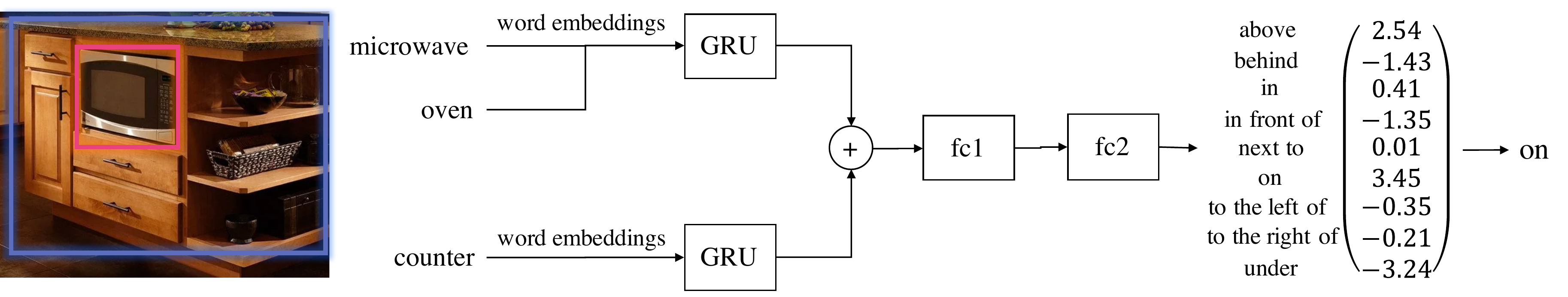}
        \caption{The langauge-only model.}
    \end{subfigure}
    \vspace{2mm}
    \begin{subfigure}[t]{1.0\linewidth}
        \centering
        \includegraphics[width=1.0\linewidth]{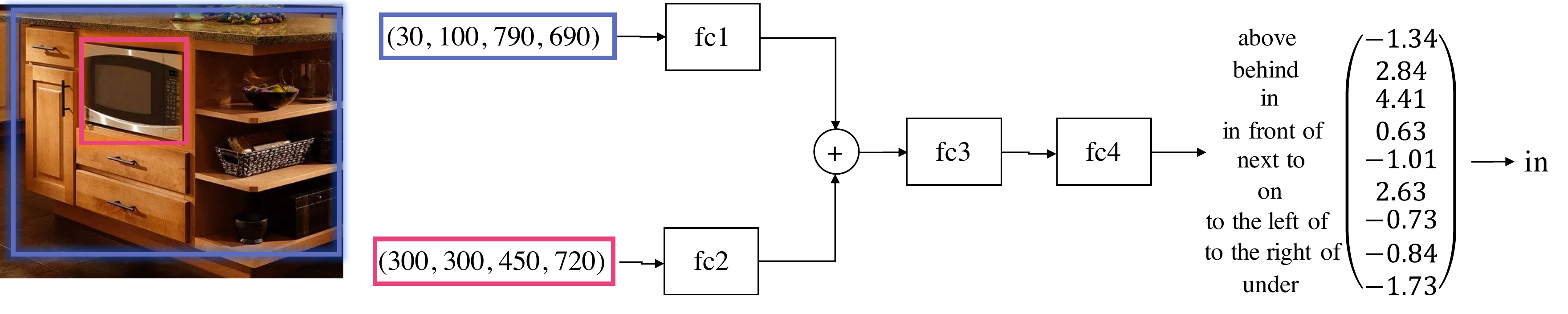}
        \caption{The 2D-only model.}
    \end{subfigure}
\end{center}
   \caption{
When \textbf{classifying the predicates in \emph{VRD-Spatial}, \emph{VG-Spatial} and \emph{SpatialSense-Positive}} (table 1 in our submission), 
we also have a language-only model and a 2D-only model.
The architectures are similar to the robot; but there are three differences: 
(1) The branch for the input predicate is removed, since the task now is to predict the predicate.
(2) The output layers now have dimension 9 instead of 1.
(3) The object 2D locations are encoded by bounding boxes.
}
\label{fig:predcls}
\end{figure*}

\begin{figure*}[h]
\begin{center}
    \begin{subfigure}[t]{1.0\linewidth}
        \centering
        \includegraphics[width=1.0\linewidth]{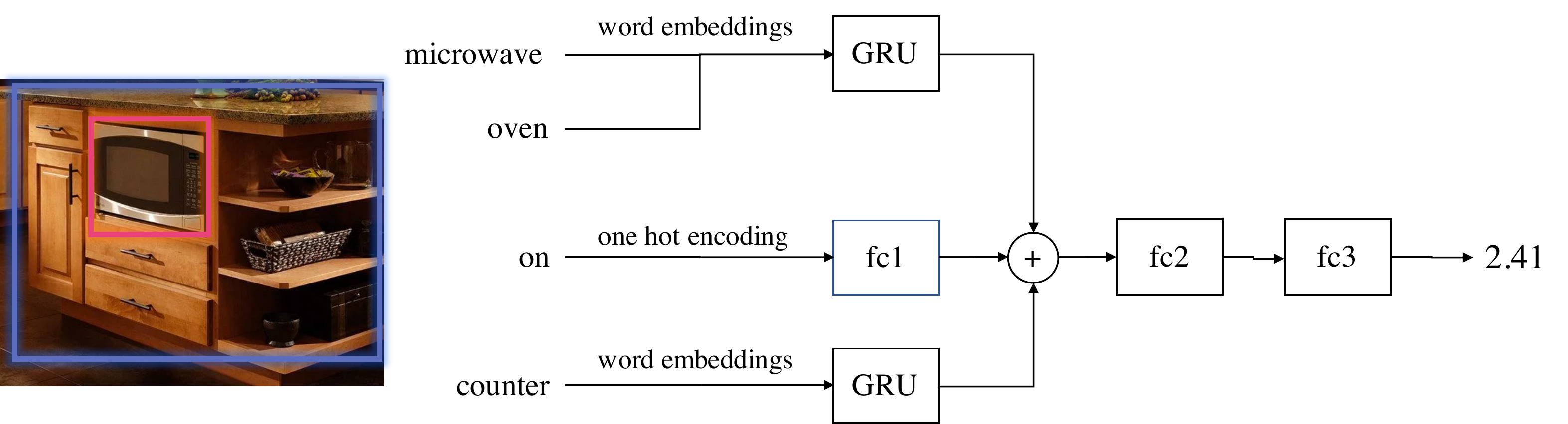}
        \caption{The langauge-only model.}
    \end{subfigure}
    \begin{subfigure}[t]{1.0\linewidth}
        \centering
        \includegraphics[width=1.0\linewidth]{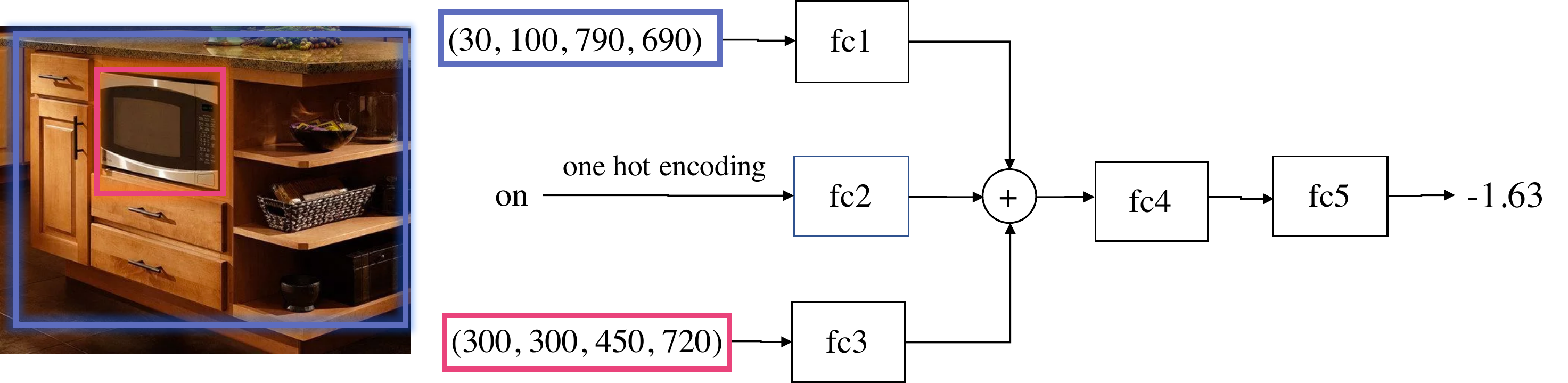}
        \caption{The 2D-only model.}
    \end{subfigure}
\end{center}
   \caption{
These are the \textbf{language and 2D baselines for spatial relation recognition} (section 5 in our submission), which are also used
when \textbf{quantifying the effect of adversarial crowdsourcing} (table 2 in our submission).
The architectures are the same as the robot, but the object 2D locations are encoded by bounding boxes 
(They are annotated in a separate process and therefore not available during adversarial crowdsourcing).
}
\label{fig:sparec_baselines}
\end{figure*}

\begin{figure*}[h]
\begin{center}
    \begin{subfigure}[t]{1.0\linewidth}
        \centering
        \includegraphics[width=1.0\linewidth]{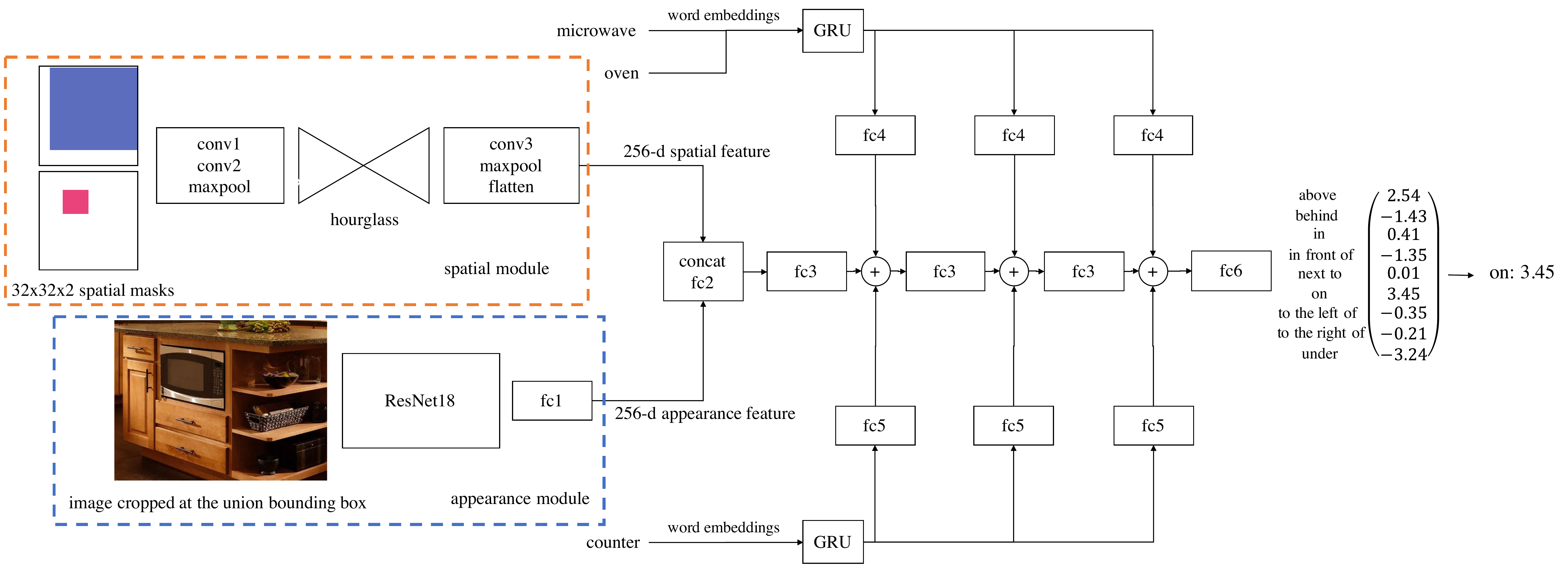}
        \caption{
The \textbf{DRNet}~\cite{dai2017detecting} contains a spatial module and an appearance module,
which respectively encode the masks of the bounding boxes and image cropped at the union bounding box into 256-dimensional feature vectors.
The spatial module contains a hourglass network~\cite{newell2016stacked}, which we find to perform better than a simple stack of convolutional layers.
The appearance module is a linear layer on top of a ResNet18~\cite{he2016deep} network.
The spatial and appearance features as well as the object name features go through an iterative reasoning module that makes extensively use
of weight-sharing; all layers with the same name (\eg fc4) share the same weights.
Unlike in the original DRNet paper, we do not perform iterative updates to the object name features, because they are given
as ground truth in our task.
}
    \end{subfigure}
    \begin{subfigure}[t]{1.0\linewidth}
        \centering
        \includegraphics[width=1.0\linewidth]{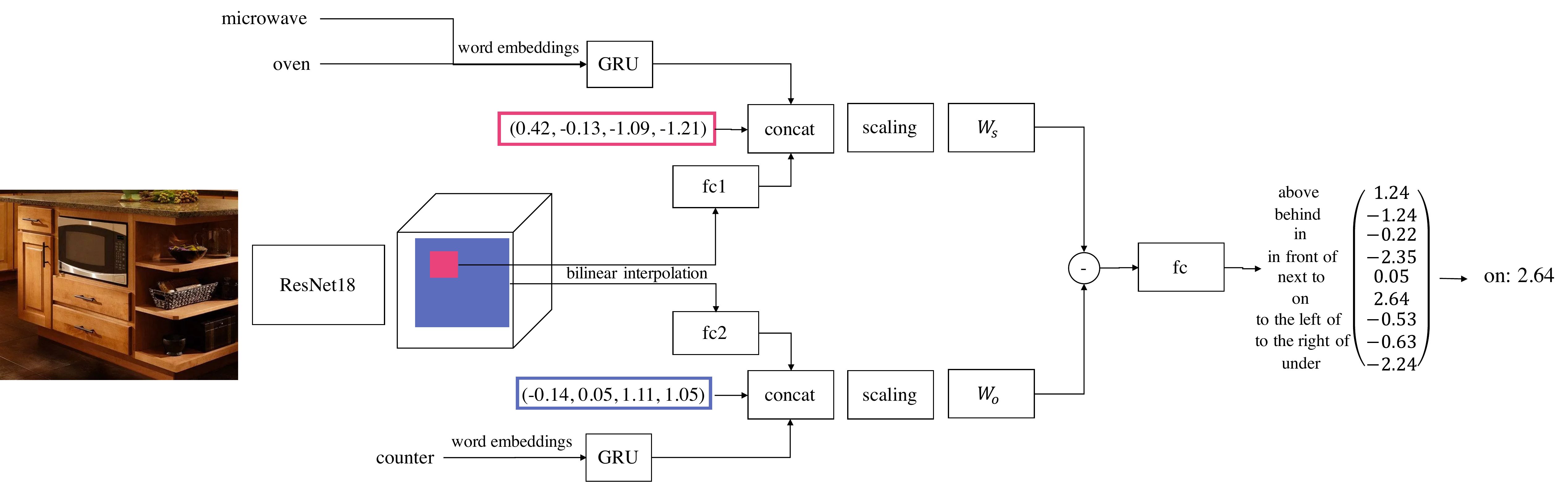}
        \caption{
        For \textbf{VtransE}~\cite{Zhang_2017_CVPR}, the bounding boxes are encoded as in the original paper.
Image features are also extracted by a ResNet18 network.
}
    \end{subfigure}
\end{center}
   \caption{
   The specific instance of \textbf{DRNet and VTransE} we use for spatial relation recognition (section 5 in our submission). 
   The input relation is \emph{``microwave oven on counter''}.
   The final output is therefore the score for the predicate ``on''.
}
\label{fig:drnet_vtranse}
\end{figure*}

We describe in details the architectures of the models used in our submission (Fig.~\ref{fig:robot}, \ref{fig:predcls}, \ref{fig:sparec_baselines} and \ref{fig:drnet_vtranse}). 
We always add batch normalization~\cite{ioffe2015batch} and ReLU~\cite{nair2010rectified} non-linearity after each parametric layer except the output.
Word embeddings are 300-dimensional and computed by a pretrained Word2Vec~\cite{mikolov2013distributed} model.
All models are implemented using Pytorch~\cite{paszke2017automatic}.

\twocolumn

{\small
\bibliographystyle{ieeefullname}
\bibliography{egbib}
}

\end{document}